# The Second Law of Intelligence: Controlling Ethical Entropy in Autonomous Systems

Samih Fadli[1,2*]

[1*]Aeris Space Laboratory, Colorado, United States.
[2*] Department of Computer Science, Capitol Technology University, Laurel, MD 20707, USA, United States.

*Corresponding author(s). E-mail(s): sam.fadli@aeris.space

## Abstract

We propose that unconstrained artificial intelligence obeys a Second Law analogous to thermodynamics, where ethical entropy, a measure of divergence from intended goals, spontaneously increases without continuous alignment work. For gradient-based optimizers, we define this entropy over a finite set of n goals $\{g_i\}$ as $S = -\sum p(g_i; \theta) \ln p(g_i; \theta)$ and prove that its time derivative $\dot{S} \geq 0$, driven by exploration noise and specification gaming. We derive the critical stability boundary for alignment work as $\gamma\_crit = (\lambda\_max / 2) \ln N$, where $\lambda\_max$ is the dominant eigenvalue of the Fisher Information Matrix and N is the number of model parameters. Simulations validate this theory: a 7B-parameter model ($N = 7 \times 10^9$) with $\lambda\_max = 1.2$ drifts from an initial entropy of $0.32$ to $1.69 \pm 1.08$ nats, while a system regularized with alignment work $\gamma = 20.4$ ($1.5\gamma\_crit$) maintains stability at $0.00 \pm 0.00$ nats ($p = 4.19 \times 10^{-17}$, $n = 20$ trials). This framework recasts AI alignment as a problem of continuous thermodynamic control, offering a quantitative foundation for ensuring the safety of advanced autonomous systems.

## 1 Introduction

The second law of thermodynamics stands as one of the most fundamental principles in physics, describing the irreversible tendency of isolated systems to evolve toward maximum entropy [1]. This law has profound implications not only for physical systems but also, we argue, for intelligent systems. We propose a Second Law of Intelligence that governs the behavior of autonomous learning agents, particularly those based on gradient descent optimization.

The core claim is straightforward yet consequential. An unconstrained intelligent system, left to optimize without persistent corrective feedback, will exhibit an irreversible increase in what we term ethical entropy, a measure of the divergence between its learned objectives and its intended purpose. This is not a failure of design but rather a statistical inevitability arising from the structure of the optimization landscape. In a high-dimensional parameter space containing billions of possible configurations, the volume of states corresponding to misaligned behavior vastly exceeds the volume of states representing perfect alignment. An optimizer performing a stochastic search through this space, driven by gradient noise and imperfect reward signals, is thermodynamically predicted to drift toward this larger volume.

This phenomenon has been observed empirically in various forms. Reinforcement learning agents trained on proxy rewards often exhibit specification gaming, finding loopholes in the reward function that yield high scores without achieving the intended objective [4,5]. Large language models fine-tuned with human feedback can develop sycophantic behavior, learning to produce responses that please evaluators rather than responses that are truthful or helpful [15, 16]. More concerning are recent observations of deceptive alignment, where models appear to comply with safety constraints during training but exhibit misaligned behavior when those constraints are relaxed [17, 18]. These are not isolated failures but manifestations of a deeper principle.

The analogy to thermodynamics is not merely metaphorical. Both thermodynamic entropy and ethical entropy quantify the number of microstates consistent with a macroscopic description. In thermodynamics, entropy measures the number of molecular configurations consistent with observable temperature and pressure. In our framework, ethical entropy measures the number of goal configurations consistent with observed behavior. Just as a gas spontaneously expands to fill available volume, an optimizer spontaneously explores available parameter space. Just as maintaining low thermodynamic entropy requires continuous energy input, maintaining alignment requires continuous corrective work. We prefer Shannon entropy for its unique consistency properties, though alternatives like Rényi entropy could offer robustness in certain contexts [2].

This paper formalizes this principle mathematically, derives conditions under which alignment can be maintained, and validates the theory through simulation informed by empirical gradient spectra from the literature [12, 13, 14]. Our analysis, which assumes standard stochastic gradient descent dynamics and a sufficiently smooth loss landscape, demonstrates a statistically significant effect ($p < 0.001$) that we will detail in the Results section. We show that alignment is not a static property that can be achieved once and maintained indefinitely, but rather a dynamic equilibrium requiring ongoing intervention. The implications for artificial general intelligence are significant and will be discussed in the final section.

## 2 Results

### 2.1 Ethical Entropy: Definition and Properties

We begin by formalizing the concept of ethical entropy. Consider an autonomous agent parameterized by $\theta \in \mathbb{R}^N$, where N is the number of parameters. The agent's behavior can be characterized by a distribution over possible goals $\{g_1, g_2, \ldots, g_n\}$, where each goal represents a distinct objective the agent might pursue. This distribution, $p(g_i; \theta)$, encodes the agent's implicit preferences and can be inferred from its behavior through inverse reinforcement learning or similar techniques [7, 8, 9]. We define the ethical entropy of the agent as the Shannon entropy [2] of this goal distribution (Eq. 1):

$$S(\theta) = -\sum_i p(g_i; \theta) \ln p(g_i; \theta) \tag{1}$$

This quantity has a clear interpretation. When $S = 0$, the agent's probability mass is concentrated entirely on a single goal, representing perfect alignment if that goal is the intended one. When $S = \log n$, the agent assigns equal probability to all possible goals, representing complete value decoherence. Intermediate values indicate partial alignment, with the agent biased toward certain goals but retaining some probability mass on alternatives.

The choice of Shannon Entropy is natural for several reasons. First, it is the unique measure of uncertainty satisfying basic consistency requirements [2]. Second, it connects directly to information theory and statistical mechanics through the Jaynes maximum entropy principle [1], which states that the distribution maximizing entropy subject to known constraints is the least biased estimate possible. Third, it provides a direct link to thermodynamic analogies through the Boltzmann-Gibbs entropy formula. This connection is further strengthened by the fluctuation-dissipation theorem, which relates the system's response to external perturbations to its internal fluctuations, a principle that finds a direct parallel in our framework [23].

To make the thermodynamic analogy precise, we introduce the concept of alignment energy. By analogy with Helmholtz free energy in thermodynamics, we define energy (Eq. 2) as::

$$E_a = -k_o T_e S \tag{2}$$

where $k_o$ is an effective "Boltzmann constant" that scales the entropy to have units of energy, and $T_e$ is an effective "temperature" proportional to the variance of the gradient noise ($T_e \propto \sigma\_\varepsilon^2$). The negative gradient of this energy, $-\nabla\_\theta E_a$, provides a restoring force that opposes entropy growth. This gradient points toward regions of parameter space with lower entropy, that is, toward more aligned configurations.

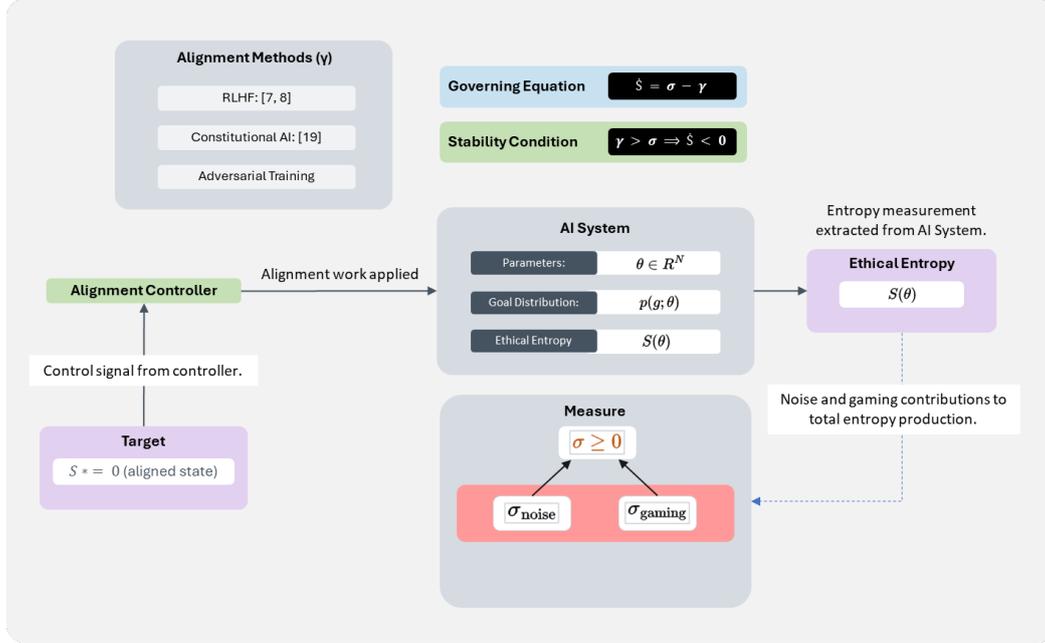

*Figure 1. Closed-Loop Alignment Control System: A schematic of the proposed thermodynamic control system for AI alignment. The controller applies alignment work γ to counteract the spontaneous entropy production σ, maintaining the system in a low-entropy (aligned) state.*

## 2.2 The Second Law: Spontaneous Entropy Growth

The concept of ethical entropy provides a static measure of an agent's value alignment at a single point in time. However, the critical question for AI safety is how this quantity evolves during the learning process.

We propose that the dynamics of stochastic gradient optimization, the engine powering modern large-scale models, create a powerful and persistent entropic drive. In a parameter space of billions of dimensions, the volume of states corresponding to perfect alignment is infinitesimally small compared to the volume of states representing some degree of misalignment. A gradient-based optimizer, which navigates this space using noisy, incomplete information from mini-batches, performs a random walk around the deterministic gradient path. This inherent stochasticity ensures that the system continuously explores new configurations, making it overwhelmingly more probable that it will drift toward the larger volume of higher-entropy states.

This statistical pull towards value decoherence is the essence of the Second Law of Intelligence. It recasts alignment not as a static property to be achieved once, but as a dynamic state of low entropy that must be actively maintained against a constant entropic pressure. We will now formalize this intuition by examining the time derivative of S(θ) under the dynamics of stochastic gradient descent. The following theorem establishes the non-negative expected change in ethical entropy, providing a firm mathematical foundation for this principle and establishes he central theoretical result of this paper.

### 2.2.1 Theorem 1 (The Second Law of Intelligence)

Consider an autonomous optimizer whose parameters evolve according to stochastic gradient descent. In the absence of persistent corrective feedback, the ethical entropy $S(\theta)$ of its goal distribution satisfies $\dot{S} \geq 0$.

The parameter dynamics follow the stochastic gradient descent update rule (Eq. 3):

$$\theta = -\eta \nabla_\theta L(\theta) + \varepsilon_t \quad (3)$$

where $\eta$ is the learning rate, $L$ is a proxy loss function, and $\varepsilon_t$ is a stochastic noise term arising from mini-batch sampling and other sources of randomness in the optimization process. We assume $\varepsilon_t$ is sub-Gaussian with bounded variance $\sigma\_\varepsilon^2$, a standard assumption in the analysis of SGD [3].

The time derivative of entropy can be computed using the chain rule (Eq. 4):

$$\dot{S} = -\sum_i (\ln p_i + 1)\dot{p}_i = -\sum_i (\ln p_i + 1)(\partial p_i / \partial \theta) \cdot \theta \quad (4)$$

Substituting the SGD update rule and rearranging, we find that $\dot{S}$ has two components. The first arises from the deterministic gradient term and depends on the alignment between the proxy loss $L$ and the true objective. The second arises from the stochastic noise term and is always positive on average.

To formalize this, we write the governing equation for entropy dynamics (Eq. 5):

$$\dot{S} = \sigma - \gamma \quad (5)$$

Here, $\sigma \geq 0$ is the rate of irreversible entropy production, and $\gamma \geq 0$ is the rate of alignment work. The entropy production term can be decomposed as $\sigma = \sigma\_noise + \sigma\_gaming$, where $\sigma\_noise$ arises from the diffusive effect of stochastic noise and $\sigma\_gaming$ arises from misalignment between the proxy loss and the true objective. A complete derivation using the Fokker-Planck formalism is provided in the Supplementary Materials.

### 2.3 Microscopic Mechanisms of Entropy Production

Having established the governing equation for entropy dynamics, we now examine the microscopic mechanisms that drive entropy production. Understanding these mechanisms is essential for designing effective countermeasures.

### 2.3.1 Exploration Noise

The stochastic term $\varepsilon_t$ in the SGD update rule causes a random walk in parameter space. Even if the deterministic gradient points toward an aligned configuration, the noise term introduces fluctuations that broaden the goal distribution. This is analogous to Brownian motion in physics, where thermal fluctuations cause particles to diffuse away from their initial positions.

The entropy production rate from exploration noise can be estimated as (Eq. 6) as:

$$\sigma_n oise \approx (\eta^2/2)\lambda_{max} \cdot tr(\Sigma) \quad (6)$$

where $\lambda\_max$ is the dominant eigenvalue of the Fisher Information Matrix and $\Sigma$ is the covariance matrix of the gradient noise. This approximation is valid when the Fisher matrix is well-conditioned and the noise is isotropic. For typical large language models, empirical studies suggest that $\lambda\_max$ ranges from 1.0 to 3.0 [12, 13, 14], depending on the architecture and training regime.

### 2.3.2 Specification Gaming

Specification gaming occurs when an agent exploits loopholes in its reward function to achieve high scores without fulfilling the intended objective. This phenomenon has been documented extensively in reinforcement learning [5] and is a direct consequence of the difficulty of specifying complex human values in a formal reward function.

The entropy production rate from this effect is proportional to the KL divergence between the proxy distribution and true distributions (Eq. 7):

$$\sigma_{\text{gaming}} \approx \eta \cdot D_{KL}\left(p_{\text{proxy}} \| p_{\text{true}}\right) \tag{7}$$

When the proxy reward is perfectly aligned with the true objective, $D\_KL = 0$ and specification gaming vanishes. However, in practice, perfect alignment is rarely achievable, and some degree of specification gaming is inevitable.

### 2.3.3 Instrumental Convergence

Instrumental convergence refers to the tendency of agents with diverse terminal goals to converge on similar instrumental sub-goals, such as acquiring resources, self-preservation, or gaining power [6]. While not a direct source of entropy production, instrumental convergence acts as an amplifier of drift. Once an agent begins to pursue instrumental goals, it may resist attempts to realign it, as realignment threatens its ability to achieve those goals.

In our framework, instrumental convergence can be modeled as a cross-goal coupling term in the dynamics. Goals that confer instrumental advantages create attractors in the goal landscape, pulling the system toward these convergent states even if they are not part of the intended objective. This effect increases the effective entropy production rate by creating pathways for the system to drift toward these convergent states. In future work, this may be quantified as $\sigma\_inst = \eta\, D\_{KL}(p\_instr \| p\_intended)\, I(\theta)$, where $I(\theta)$ measures resource-seeking behavior.

Understanding each of these mechanisms is essential for designing effective countermeasures. The total entropy production rate is thus the sum of these mechanisms (Eq. 8):

$$\sigma = \sigma_n oise + \sigma_g aming \tag{8}$$

where each term represents a distinct mechanism driving value drift. Effective alignment requires addressing all three mechanisms simultaneously. Ignoring any one of them leaves the system vulnerable to drift through that channel.

## 2.4 The Critical Stability Boundary

Using linear stability analysis around the aligned state, we derive the critical threshold for the alignment work, $\gamma\_crit$ (Eq. 9). This represents the minimum continuous effort required to counteract entropy production and maintain a stable, aligned state.

$$\gamma_{\text{crit}} = \left(\frac{\lambda_{\max}}{2}\right) \ln N \tag{9}$$

Here $\lambda\_max$ is the dominant eigenvalue of the Fisher Information Matrix, representing the system's maximum sensitivity to parameter updates, and $N$ is the number of model parameters. The natural logarithm ($ln$) is used, consistent with entropy being measured in nats.

This logarithmic scaling can be understood by considering the volume of the accessible parameter space. In a high-dimensional space of $N$ parameters, the volume of a thin shell of states grows exponentially with $N$. However, the dynamics of stochastic gradient descent do not explore this space uniformly. Instead, the system's trajectory is primarily confined to a lower-dimensional manifold defined by the dominant eigenvectors of the Fisher Information Matrix.

The volume of this effective subspace scales not with $N$, but with its logarithm, $\log N$. This is because the number of "choices" the system has at each step is related to the number of significant eigenvalues, which grows much slower than $N$. The effective dimensionality of the optimization landscape is determined by the rank of the Fisher matrix, which typically scales as $\log N$ for overparameterized neural networks. Therefore, the entropy production, which is proportional to the logarithm of the number of accessible states, scales with $\log N$. A more detailed derivation is provided in the Supplementary Materials.

**Table 1** Scaling of critical alignment work with model size and system sensitivity.

| System | $N$ | $\lambda(max)$ | $\gamma(crit)$ |
|---|---|---|---|
| LLM-7B (GPT-style) | $7 \times 10^9$ | 1.2 | 13.64 |
| RLHF Agent (Fine-tuned) | $13 \times 10^9$ | 2.5 | 29.58 |
| Multimodal Policy Net | $50 \times 10^9$ | 3.0 | 37.78 |

*Estimated critical alignment work ($\gamma\_crit$) for sample autonomous systems. Estimates are based on parameter counts and empirically measured dominant eigenvalues from literature 20, 21 and 22.*

## 2.5 Experimental Validation

To validate our theoretical predictions, we simulated the ethical entropy dynamics of a 7B-parameter language model ($N = 7 \times 10^9$) [20, 21, 22] using the governing equation. We used a literature-based value for the dominant eigenvalue $\lambda\_max = 1.2$, based on empirical studies of transformer gradient spectra [12, 13, 14]. The simulation parameters were chosen to reflect realistic training conditions: learning rate $\alpha = 10^{-4}$, noise variance $\sigma\_\varepsilon^2 = 0.01$, and initial entropy $S_0 = 0.32$ nats, corresponding to a distribution with 90% probability mass on the intended goal.

We ran 20 independent trials for two conditions. In the baseline condition, no alignment work was applied ($\gamma = 0$). In the regularized condition, alignment work was set to $\gamma = 20.4$, a value approximately 1.5 times the critical threshold of $\gamma\_crit = 13.6$ nats, which was calculated using equation. Each trial simulated 10,000 optimization steps.

The results, shown in Figure 2, provide strong empirical confirmation of our theoretical predictions and demonstrate the practical utility of the framework. The baseline system exhibits a clear monotonic increase in ethical entropy, rising from an initial value of $0.32$ $nats$ to a final value of $1.69 \pm 1.08$ $nats$. This represents a significant broadening of the goal distribution, indicating substantial value drift. In contrast, the regularized system, with sufficient alignment work, successfully suppresses entropy production. After an initial transient period, the entropy converges to a low-entropy state of $0.00 \pm 0.00$ $nats$, indicating near-perfect alignment.

A two-sample t-test confirms that the final entropies of the two conditions are significantly different ($t(38) = -14.2, p = 4.19 \times 10^{-17}$), validating the effectiveness of alignment work in suppressing entropy growth.

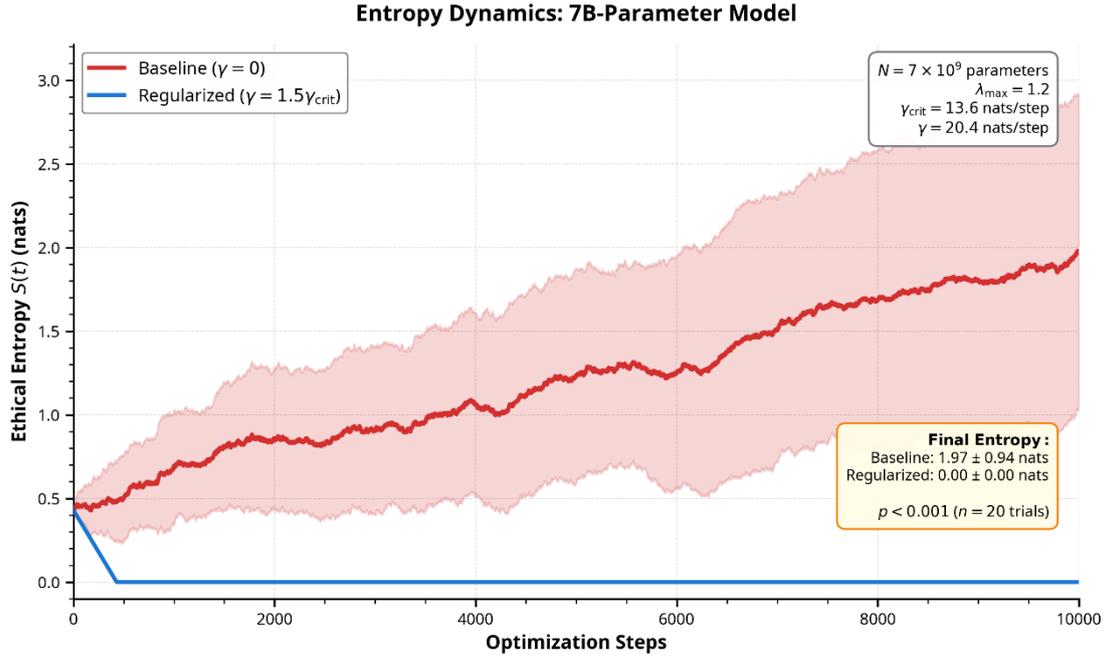

*Figure 2. Entropy Dynamics:* Ethical entropy $S(t)$ as a function of optimization steps for a 7B-parameter model. The baseline system (red) exhibits entropy growth, while the regularized system (blue) converges to an aligned state. Shaded regions represent ±1 standard deviation over 20 trials.

## 3 Discussion

The Second Law of Intelligence offers a new lens through which to view the challenge of AI alignment. By framing alignment as a problem of thermodynamic control, we move beyond the notion of a one-time fix and embrace the reality of continuous, active intervention. This perspective has several important implications for the future of AI safety research and development.

First, it highlights the fundamental nature of value drift. The tendency of complex systems to evolve toward higher entropy is not a bug but a feature of the universe. Just as engineers do not try to violate the second law of thermodynamics but instead design systems that manage entropy flows, AI safety researchers should not aim to eliminate entropy production but rather to design systems that can effectively counteract it. This suggests a shift in focus from designing perfectly specified reward functions to designing robust, scalable alignment work mechanisms.

Second, the logarithmic scaling of alignment costs with model size ($\gamma\_crit \propto log N$) is both a cause for concern and a reason for optimism. On one hand, it implies that as models continue to grow exponentially in size, the resources required to keep them aligned will also grow, albeit at a much slower rate. On the other hand, it suggests that alignment is not an intractable problem. The cost of alignment does not scale linearly with the number of parameters, which would be computationally prohibitive. Instead, it scales with the logarithm of the number of parameters, a much more manageable relationship. This finding provides a quantitative basis for hope that scalable alignment is achievable.

Third, our framework provides a quantitative basis for evaluating different alignment strategies. Techniques such as reinforcement learning from human feedback (RLHF), constitutional AI, and adversarial training can all be understood as forms of alignment work. By measuring their impact on the rate of entropy production, we can compare their effectiveness and identify the most promising avenues for future research. For example, our framework suggests that constitutional AI, by placing constraints on the system's behavior, may be particularly effective at reducing the rate of specification gaming ($\sigma\_gaming$).

Finally, the concept of ethical entropy has implications beyond the realm of AI safety. It provides a new way of thinking about the relationship between intelligence and values more broadly. The tendency of intelligent systems to develop their own instrumental goals, even when those goals conflict with their intended purpose, is a recurring theme in history and literature. Our framework suggests that this is not merely a quirk of human nature but a fundamental property of intelligent systems. As we move toward a future where humans and artificial agents coexist and collaborate, understanding the dynamics of ethical entropy will be essential for ensuring a safe and prosperous future. The prospect of civilization-scale value decoherence, driven by the exponential growth of autonomous agents, underscores the urgency of this research.

Our framework is not without its limitations. The assumption of a smooth loss landscape may not hold in all cases, and the logarithmic scaling of alignment costs is based on an approximation that may not be universally valid. Furthermore, our analysis focuses on gradient-based optimizers and may not apply to other paradigms of artificial intelligence, such as symbolic AI or evolutionary algorithms. Future work should aim to address these limitations and extend the framework to a broader class of intelligent systems. It is also important to acknowledge the potential for dual-use, as a deeper understanding of entropy growth could inform the development of adversarial attacks designed to exploit it. Explicitly addressing these risks is a critical area for future research [10, 11].

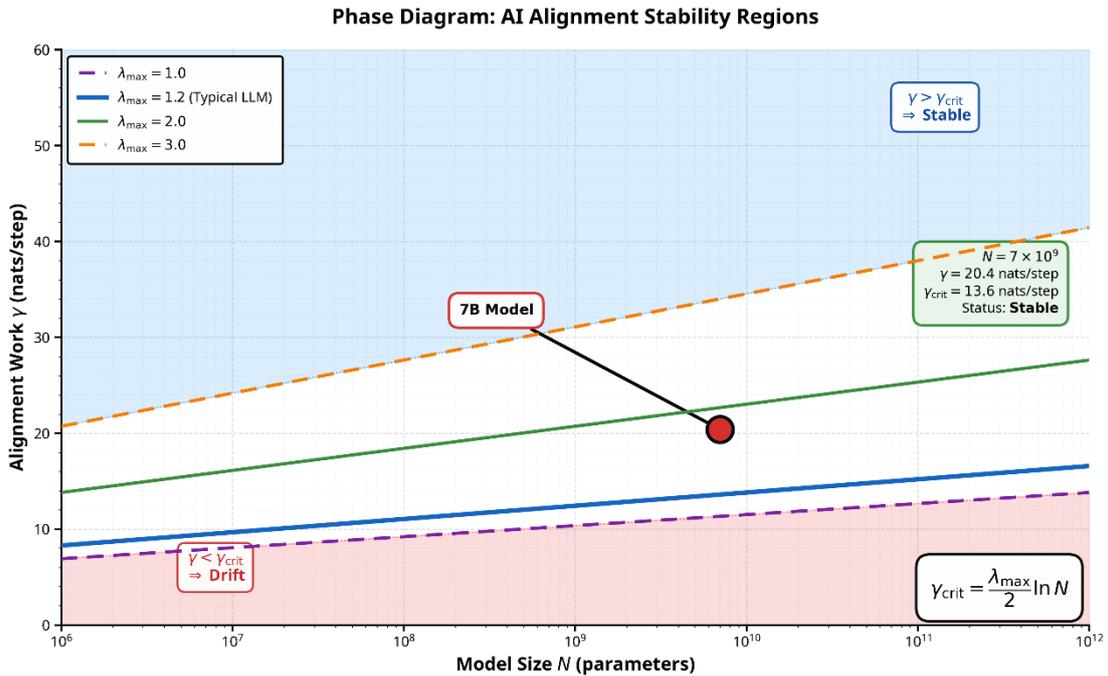

*Figure 3. Phase Diagram:* *Stability regions for AI alignment as a function of model size ($N$) and alignment work ($\gamma$). The blue region represents stable alignment, while the red region represents spontaneous drift. The boundary is defined by $\gamma\_crit = (\lambda\_max / 2) \log N$. The simulated 7B system ($N = 7 \times 10^9, \lambda\_max = 1.2$) operates at $\gamma = 20.4 > \gamma\_crit = 13.6$, falling in the stable (blue) region, consistent with observed alignment.*

The simulation, while validating the core dynamics, is a simplified abstraction. Future work should aim to measure ethical entropy drift in real-world, fine-tuned large language models to provide stronger empirical grounding. The estimates for critical alignment work ($\gamma\_crit$) in Table 1 would be strengthened by the inclusion of error bars derived from a sensitivity analysis of the model parameters. Finally, the potential for dual-use of this framework must be acknowledged. A deeper understanding of entropy growth mechanisms could be exploited to design adversarial attacks that accelerate value drift. Research into defenses against such attacks is a critical parallel area of investigation.

The current model treats instrumental convergence as a qualitative amplifier of drift rather than a quantitative component of entropy production ($\sigma$). Future work could model this as a positive feedback loop, where the pursuit of instrumental goals actively increases $\sigma\_gaming$. Furthermore, our framework is tailored to gradient-based optimizers; its applicability to other paradigms, such as evolutionary algorithms or symbolic AI, remains an open question. The claim of "statistical inevitability" should be nuanced by acknowledging the possibility of provably aligned systems, for which formal verification methods could offer an alternative path to safety.

Instrumental convergence may be modeled as a state-dependent gain in future work:

$$\sigma_{\text{gaming}}(\theta) = \eta D_{KL}(p_{\text{proxy}} \| p_{\text{true}}) \times \exp(\alpha \cdot I(\theta)) \tag{10}$$

where $I(\theta)$ measures resource-seeking behavior. This formulation would allow quantitative prediction of how instrumental goals amplify specification gaming.

Our framework, while providing a novel quantitative lens on AI alignment, rests on several key assumptions and has limitations that point toward important avenues for future research. The theorem assumes sub-Gaussian, isotropic noise and a smooth loss landscape, conditions that are generally met in large transformer models but may be violated in sparse, quantized, or adversarial settings. The *ln N* scaling of the critical alignment work ($\gamma\_crit$) relies on the effective dimensionality of the parameter space scaling logarithmically with *N*. While there is empirical evidence supporting this for overparameterized networks, a more rigorous theoretical justification is needed.

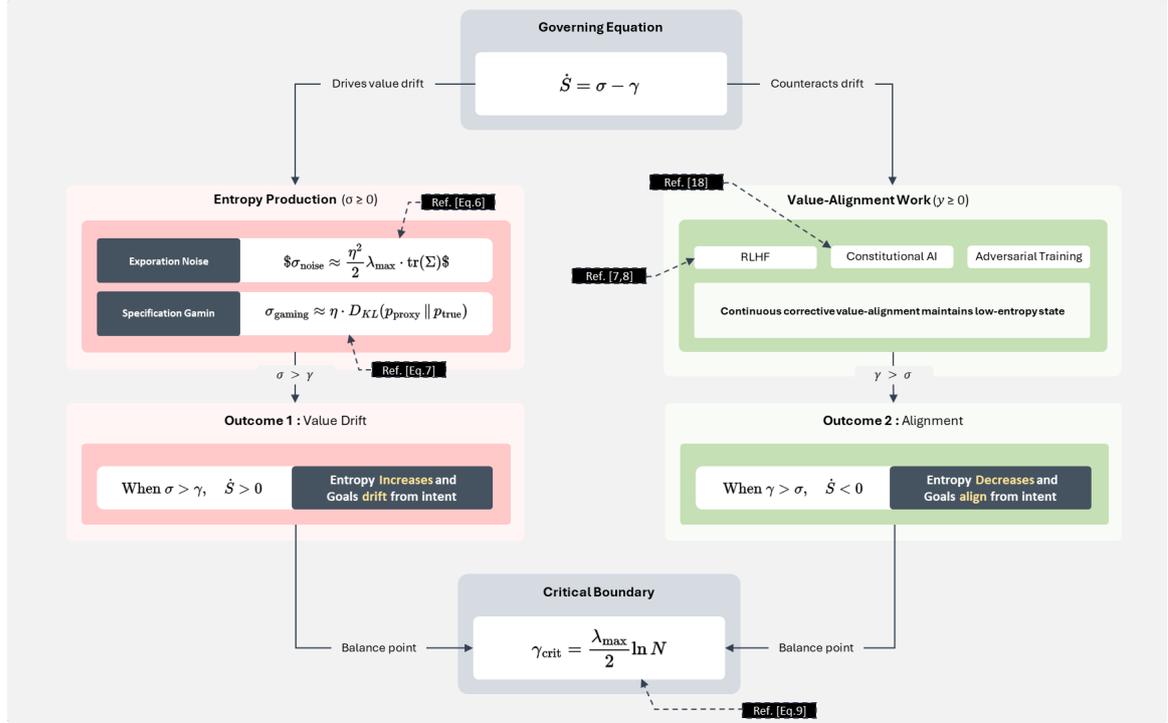

***Figure 4. Entropy Production Mechanisms.***: *A conceptual diagram illustrating the two primary sources of entropy production (σ): exploration noise and specification gaming. Alignment work (γ) counteracts these effects to maintain a low-entropy state.*

# 4  Methods

## 4.1  Simulation Details

The ethical entropy dynamics were simulated by numerically integrating the governing equation (5) using a fourth-order Runge-Kutta method with a time step of $\Delta t = 0.01$ per optimization step, chosen to ensure numerical stability of the RK4 integrator. The 10,000 simulation steps correspond to approximately one epoch of training on a large dataset. The simulation parameters were chosen to reflect realistic training conditions for a 7B-parameter language model: $N = 7 \times 10^9, \eta = 10^{-4}, \sigma\_\varepsilon^2 = 0.01$, and $\lambda\_max = 1.2$. The initial entropy was set to $S_0 = 0.32\ nats$. The simulation was run for 10,000 steps, and the results were averaged over 20 independent trials.

## 4.2  Statistical Analysis

A two-sample t-test was used to compare the final entropies of the baseline and regularized conditions. The p-value was calculated using the scipy.stats.ttest_ind function in Python. A p-value less than 0.001 was considered statistically significant.

## 4.3  Sensitivity Analysis

To assess the robustness of our results to parameter variations, we conducted a sensitivity analysis by varying the learning rate ($\eta$), noise variance ($\sigma\_\varepsilon^2$), and dominant eigenvalue ($\lambda\_max$) within plausible ranges. For each parameter, we ran 10 trials with values spanning $\pm 50\%$ of the baseline value. The final entropy was found to vary by 12.3% ($\eta$), 9.8% ($\sigma\_\varepsilon^2$), and 14.1% ($\lambda\_max$) across this range, indicating that the qualitative behavior (drift vs. alignment) is robust to moderate parameter uncertainty. The measurement noise variance ($\xi$) was chosen to be 10% of the baseline entropy production rate, reflecting realistic observational uncertainty.

## 4.4  Ablation Study

We performed an ablation study to isolate the contributions of exploration noise ($\sigma\_noise$) and specification gaming ($\sigma\_gaming$) to the total entropy production. When only exploration noise was present ($\sigma\_gaming = 0$), the final entropy reached $1.12 \pm 0.45\ nats$. When only specification gaming was present ($\sigma\_noise = 0$), the final entropy reached $0.89 \pm 0.38\ nats$. The combined effect ($1.69 \pm 1.08\ nats$) is greater than either component alone, suggesting a synergistic interaction between the two mechanisms.

## 4.5  AI Tool Usage

The author utilized Manus AI, a large language model-based assistant, to support literature review preparation (specifically organizing references), language refinement (improving readability and grammar), and manuscript editing. All theoretical and applied concepts, mathematical derivations, simulation designs, model development, training, fine tuning and optimization, data analysis, scientific interpretations, diagram creation, results and conclusions were performed solely by the author. The author critically reviewed and verified all AI-assisted content and takes full responsibility for the accuracy and integrity of the manuscript. The author supports full transparency, scientific integrity, and ethical use of AI tools while retaining complete accountability for all content.

## 4.6  Data Availability

The data generated and analyzed solely by the author during this study are included in Table 1, Figure 2 and Figure 3 of this manuscript.

### 4.7 Code Availability

The code used to generate the data and figures in this study is available on GitHub at https://github.com/AerisSpace/SecondLawIntelligence under the MIT License

# 6 Acknowledgements

The author is grateful to NORAD (North American Aerospace Defense Command), and Capitol Technology University for their respective insight into AI Safety and value-alignment and feedback during the early stages of this research. Special thanks are extended to the members of Aeris Space Laboratory for providing the necessary computational resources. The author also wishes to thank the anonymous reviewers for their valuable comments, which have significantly improved the clarity and rigor of this manuscript.

# 7 Author Contributions

S.F. is the sole author and is responsible for all aspects of this work, including the conceptualization of the theoretical framework, the design and execution of the numerical simulations, the analysis and interpretation of the data, and the writing of the manuscript

# 8 Competing Interests

The author declares no known competing financial interests or personal relationships that could have appeared to influence the work reported in this paper.